\documentclass[12pt]{elsarticle}

\usepackage{amssymb}
\usepackage{subcaption}
\usepackage{multirow}
\usepackage{makecell}
\usepackage{amsmath}
\usepackage{afterpage}
\usepackage{rotating}
\usepackage{subcaption}
\usepackage{algorithm}
\usepackage{lineno}
\usepackage{algpseudocode}
\usepackage{amsmath}
\usepackage{graphicx}
\usepackage{adjustbox}

\usepackage[utf8]{inputenc} 
\usepackage[T1]{fontenc}    
\usepackage{hyperref}       
\usepackage{url}            
\usepackage{booktabs}       
\usepackage{amsfonts}       
\usepackage{nicefrac}       
\usepackage{microtype}      
\usepackage{lipsum}
\usepackage{graphicx}
\graphicspath{{media/}}     
\usepackage{float}

\journal{arxiv}

\begin{document}

\begin{frontmatter}



\title{BEYONDWORDS is All You Need: Agentic Generative AI based Social Media Themes Extractor}


\author[label1]{Mohammed-Khalil Ghali} 
\author[label1]{Abdelrahman Farrag} 
\author[label1]{Sarah Lam}
\author[label1]{Daehan Won}


\affiliation[label1]{organization={School of Systems Science and Industrial Engineering, State University of New York at Binghamton},
            addressline={4400 Vestal Pkwy}, 
            city={Binghamton},
            postcode={13902}, 
            state={NY},
            country={USA}}

\begin{abstract}
Thematic analysis of social media posts provides a major understanding of public discourse, yet traditional methods often struggle to capture the complexity and nuance of unstructured, large-scale text data. This study introduces a novel methodology for thematic analysis that integrates tweet embeddings from pre-trained language models, dimensionality reduction using  and matrix factorization, and generative AI to identify and refine latent themes. Our approach clusters compressed tweet representations and employs generative AI to extract and articulate themes through an agentic Chain of Thought (CoT) prompting, with a secondary LLM for quality assurance.
This methodology is applied to tweets from the autistic community, a group that increasingly uses social media to discuss their experiences and challenges. By automating the thematic extraction process, the aim is to uncover key insights while maintaining the richness of the original discourse. This autism case study demonstrates the utility of the proposed approach in improving thematic analysis of social media data, offering a scalable and adaptable framework that can be applied to diverse contexts. The results highlight the potential of combining machine learning and Generative AI to enhance the depth and accuracy of theme identification in online communities.
\end{abstract}


\begin{highlights}

\item The paper presents BEYONDWORDS, a novel generative AI-based framework for theme extraction, aiming to automatically extract and refine thematic structures from large-scale social media data
\item It integrates embeddings, autoencoder neural network for dimensionality compression, matrix factorization, and Chain of Thought prompting with Large Language Models(LLMS) to recursively uncover and enhance the themes
\item The results show that the three main topics that are widely discussed in autistic community tweets are \textbf{social media content quality and engagement; advocacy for autistic rights and acceptance; mental health and well-being}
\end{highlights}


\end{frontmatter}



\section{Introduction}
The rapid growth of social media has created vast repositories of user-generated content, offering unique insights into diverse communities and social phenomena. For researchers, social media platforms such as X(Twitter previously) provide rich datasets to study public discourse, opinions, and trends. However, the unstructured and high dimensional nature of social media data presents challenges for thematic analysis, as traditional methods, like manual coding or frequency-based approaches (e.g., TF-IDF), often fail to capture the semantic depth and context embedded in posts. These limitations call for more sophisticated techniques capable of handling large-scale textual data and uncovering latent themes that reflect the nuanced perspectives expressed online.

In recent years, advancements in natural language processing (NLP) have introduced embedding techniques powered by pre-trained language models, which enable the transformation of raw text into dense, high dimensional vector representations that encapsulate both semantic meaning and contextual information. These embeddings, when coupled with dimensionality compression techniques such as autoencoder neural network and matrix factorization, can be leveraged to identify underlying themes in vast corpora of social media posts. Furthermore, the integration of generative AI models offers a novel approach to extracting and articulating these themes, moving beyond surface-level keyword extraction to generate coherent and meaningful and accurate themes.

This paper presents a general methodology for thematic analysis of social media data that combines embedding-based representations, dimensionality reduction, and generative AI. The proposed framework consists of three key components: (1) extracting social media embeddings using pre-trained language models, (2) applying dimensionality compression through autoencoder neural network, matrix factorization to uncover latent themes, and (3) utilizing generative AI through LLMs with a Chain of Thought (CoT) prompting mechanism to extract and refine these themes. 

As a case study, this methodology is applied to the analysis of tweets from the autistic community. Social media platforms have become an important space for autistic individuals to share their experiences, challenges, and perspectives on autism, often bypassing traditional gatekeepers of discourse. However, manual thematic analysis of such posts is labor-intensive and subject to biases introduced by human coders. By automating the process through the integration of machine learning and AI techniques, this methodology aims to extract themes that offer a more comprehensive and nuanced understanding of the discourse within the autistic community.

The contributions of this paper are twofold. First, it proposes a scalable and adaptable framework for thematic analysis of large-scale social media datasets, addressing the limitations of existing methods. Second, it provides insights into the experiences and challenges of the autistic community, potentially supporting more specialized care for this community, which requires the utmost support. This case study demonstrates the practical applicability of the methodology. The approach can be adapted for thematic analysis in other contexts, making it a versatile framework for researchers studying online discourse across various social media platforms.

The paper is structured as follows: Section 2 reviews related work on social media analysis, Section 3 discusses the components of the proposed methodology in detail, Section 4 discusses the case study and the dataset used for testing the proposed framework and Section 5 presents the results of the case study. Section 6 concludes with an evaluation of the strengths and limitations of the approach and a discussion of future directions for research.

\section{Related work}\label{sec:LR}

Thematic analysis is a key qualitative method used to identify patterns and themes within textual data. As social media has become a rich source of public discourse, the ability to analyze this unstructured and large-scale data has become crucial. Traditional methods of thematic analysis, though valuable for small datasets, often struggle with the volume, diversity, and complexity of social media text. This literature review outlines the progression of thematic analysis methods, highlighting the limitations of traditional techniques and how recent advancements, particularly our proposed approach, fill these gaps.
\newline
\subsection{Traditional Thematic Analysis in Social Media Research}

Thematic analysis has long been used to explore qualitative data, with foundational guidelines established by \cite{braun2006using}. Their method involves a systematic, manual coding process that identifies recurring themes through a close reading of the text. While this method remains highly respected in qualitative research, its application to social media data is problematic due to the large volume and complexity of text involved \cite{braun2019reflecting}.
Manual coding becomes time-consuming and unfeasible when applied to datasets that may include millions of posts, such as Twitter data \cite{boyatzis1998transforming}. In such cases, traditional thematic analysis often requires a reduced sample size or relies on researchers' subjective judgments to identify salient themes. This introduces biases and limits the generalizability of the findings \cite{markham2017research}. Additionally, the informal, fast-paced nature of social media language—frequently composed of abbreviations, slang, and emotive content—makes manual coding less reliable, as it struggles to capture nuances or variations in context \cite{golder2014digital}.
Due to these limitations, researchers began exploring computational methods for thematic analysis. Early approaches like word frequency analysis and topic modeling with Latent Dirichlet Allocation (LDA) became popular tools to analyze large datasets \cite{blei2003latent}. LDA has been applied extensively in social media research, particularly in studies focused on public opinion, social movements, and political discourse \cite{nguyen2015sentiment}. However, LDA's assumption that topics are merely distributions of words across documents oversimplifies the language and cannot capture the nuances of meaning, sentiment, and context present in complex social media texts \cite{chuang2012termite}.
Moreover, topic modeling techniques such as LDA struggle with the diverse, informal nature of social media language, often resulting in themes that are overly broad or too fragmented to be meaningful \cite{hong2010empirical}. Our proposed methodology directly addresses these limitations by incorporating advanced techniques like tweet embeddings from pre-trained language models, enabling a more nuanced understanding of social media discourse by preserving contextual relationships between words \cite{devlin2018bert}.
\newline
\subsection{Advances in Machine Learning and NLP for Thematic Analysis}

Recent advancements in natural language processing (NLP) have enhanced thematic analysis techniques, allowing researchers to overcome some of the limitations posed by traditional methods. The introduction of transformer-based language models, such as BERT \cite{devlin2018bert}, GPT-2/3 \cite{brown2020language}, and others, has significantly improved the ability to understand semantic relationships in text. These models leverage vast amounts of training data to generate contextualized word embeddings, offering deeper insights into text than earlier models like LDA or TF-IDF \cite{mikolov2013distributed}.

Two main studies rely on topic modeling and sentiment analysis to understand themes related to autism on Twitter. The first study by \cite{corti2022social} uses Non-Negative Matrix Factorization (NMF) for topic modeling and sentiment analysis with AFiNN, while the second study \cite{gabarron2023autistic} applies the BERTopic model for clustering tweets and extracting themes using dimensionality reduction via UMAP. Despite their contributions, these approaches exhibit limitations. NMF, although useful for topic coherence, lacks the ability to capture deep linguistic nuances, particularly in social media language. The BERTopic approach, while more advanced, relies on static embeddings and bag-of-words methods that may overlook semantic richness and context within tweets, especially in complex discourse like the autism community’s.

The use of pre-trained language models in thematic analysis has been explored in recent studies. \cite{wu2022study} applied BERT embeddings to cluster social media posts about public health issues, demonstrating improved coherence in the resulting themes compared to LDA. Likewise, \cite{yin2020detecting} showed that using contextualized embeddings from language models improved the ability to detect latent topics in crisis communication data. These studies highlight the superior capacity of pre-trained models to handle social media’s evolving terminology and informal language structures.

However, while embeddings capture rich semantic information, high dimensional representations pose challenges for downstream analysis. Dimensionality reduction techniques like autoencoder neural network and matrix factorization have emerged as effective ways to reduce complexity while preserving thematic content. \cite{chauhan2024tracking} demonstrated the utility of autoencoder neural network in compressing embeddings for more efficient clustering in sentiment analysis, which improves both the scalability and interpretability of results.

Our methodology builds on these approaches by integrating autoencoder neural network with matrix factorization to uncover latent themes, retaining the rich semantic information encoded in the tweet embeddings while effectively reducing dimensionality. This combination ensures that nuanced themes can be extracted from large datasets without losing the contextual meaning captured by the language models.
\newline
\subsection{Generative AI and Iterative Theme Refinement}

Generative AI models, such as GPT-3, have shown promise in thematic analysis by automating tasks such as text summarization, classification, and thematic extraction \cite{brown2020language}. These models generate coherent and contextually appropriate text, which can be useful for extracting or refining themes from large datasets. For example, studies by \cite{dong2024understanding} explored generative models for summarizing social media content, underscoring their potential for condensing vast amounts of text into coherent themes.

Despite this potential, generative AI has rarely been used in a more iterative, refining role in thematic analysis. Previous studies have typically applied generative models for summarization or categorization, without refining or validating the themes through multiple stages of analysis. This paper's \cite{wanna2024topictag} use CoT prompting is limited in that it primarily focuses on improving topic labels through token-based features and manual filtering, without fully leveraging the iterative reasoning potential of CoT for deeper semantic understanding. Additionally, their approach to CoT is more narrowly applied to optimize prompt tuning rather than refining the thematic structure of the topics themselves.
This gap is addressed in our methodology by using an iterative CoT prompting mechanism, which allows for iterative theme extraction and refinement. This process ensures that the themes are coherent, contextually accurate, and relevant to the dataset.

Although recent advancements in NLP and machine learning have improved thematic analysis, several critical gaps remain. First, many existing models still struggle to capture the complexity and nuance of informal social media language, particularly in communities with specialized or evolving vocabularies, such as the autistic community. LDA and word frequency-based methods, while useful for structured data, oversimplify language and miss deeper contextual meanings \cite{chuang2012termite}.
Second, many machine learning methods lack iterative processes for refining and validating themes. Pre-trained language models and clustering methods have advanced theme extraction, but they do not inherently include mechanisms for ensuring thematic consistency or addressing ambiguous themes \cite{wu2022study,yin2020detecting}. Moreover, generative AI models applied to thematic analysis have mostly focused on single-pass theme identification without multiple stages of validation or refinement, leading to incomplete or inconsistent results and the LLMs have fixed context windows \cite{dong2024understanding} and are not able to ingest text beyond that which is why clustering needs to be done before feeding the results to LLMs.
The proposed methodology addresses these gaps by:
\begin{enumerate}
    \item Generating tweet embeddings from pre-trained language models to capture semantic relationships and nuances in social media language. This is crucial for analyzing discourse in specialized communities, like the autistic community, which is the focus of this case study.
    \item Applying dimensionality reduction by integrating autoencoder neural network and matrix factorization to reduce the complexity of embeddings without losing key thematic information while clustering the relevant social media posts together. This approach overcomes the limitations of high-dimensional data.
    \item To balance the high cost of token generation using LLMs with the need to retain meaningful cluster information, a sample size was chosen that is representative enough to capture the overall themes of the entire cluster while minimizing the data passed to the LLMs.
    \item Using generative AI for iterative refinement by employing agentic CoT prompting mechanism to iteratively extract and refine themes of the extracted clusters, ensuring that the identified themes are accurate and contextually relevant. The inclusion of a secondary LLM for quality control further enhances the reliability of the analysis, addressing common issues with theme consistency in automated methods.
\end{enumerate}
Table \ref{tab:lit_rev} summarizes previous research conducted to address the problem of social media theme extraction and highlights how the proposed methodology aims to fill the gaps identified in these studies.
\begin{table}[H]
\centering
\caption{Comparison of Thematic Analysis Methodologies for Social Media Research}
\label{tab:lit_rev}
\begin{adjustbox}{max width=\textwidth}
\begin{tabular}{>{\raggedright\arraybackslash}p{3.5cm} >{\raggedright\arraybackslash}p{3.5cm} >{\raggedright\arraybackslash}p{3.5cm} >{\raggedright\arraybackslash}p{3.5cm}}
\hline
\textbf{Methodology} & \textbf{Focus} & \textbf{Weaknesses} & \textbf{Advantage of Proposed Methodology} \\ \hline
Traditional Thematic Analysis \cite{braun2006using} & Manual coding of text  & Unfeasible for large datasets, prone to bias & Scalable embedding-based approach retains theme detail while handling large volumes \\ \hline
Latent Dirichlet Allocation (LDA) \cite{blei2003latent} & Probabilistic topic modeling for large datasets  & Overly broad themes, lacks nuanced context & Embedding-based clustering captures semantic nuance; maintains coherence \\ \hline
Non-Negative Matrix Factorization (NMF) \cite{corti2022social} & Topic coherence via dimensionality reduction  & Limited in capturing informal language nuance & Embeddings and autoencoder neural network refine themes with more linguistic context \\ \hline
BERTopic \cite{gabarron2023autistic} & Clustering and topic modeling using static embeddings & Loses contextual depth in dynamic social media language & BERT embeddings with CoT refine themes; more adaptable to language dynamics \\ \hline
Pre-trained Language Models (e.g., BERT) \cite{devlin2018bert} & Contextualized embeddings for semantic insights & High-dimensional output challenging for analysis & Dimensionality reduction (autoencoder) makes clustering efficient, preserves semantic richness \\ \hline
Generative AI \cite{dong2024understanding} & Theme extraction via summarization & Single-pass extraction; lacks iterative refinement & CoT prompting iteratively refines themes, ensures context relevance \\ \hline
\end{tabular}
\end{adjustbox}
\end{table}

\section{Methodology}\label{sec:method}
\begin{figure}[H]
	\centering
	\includegraphics[width=\textwidth]{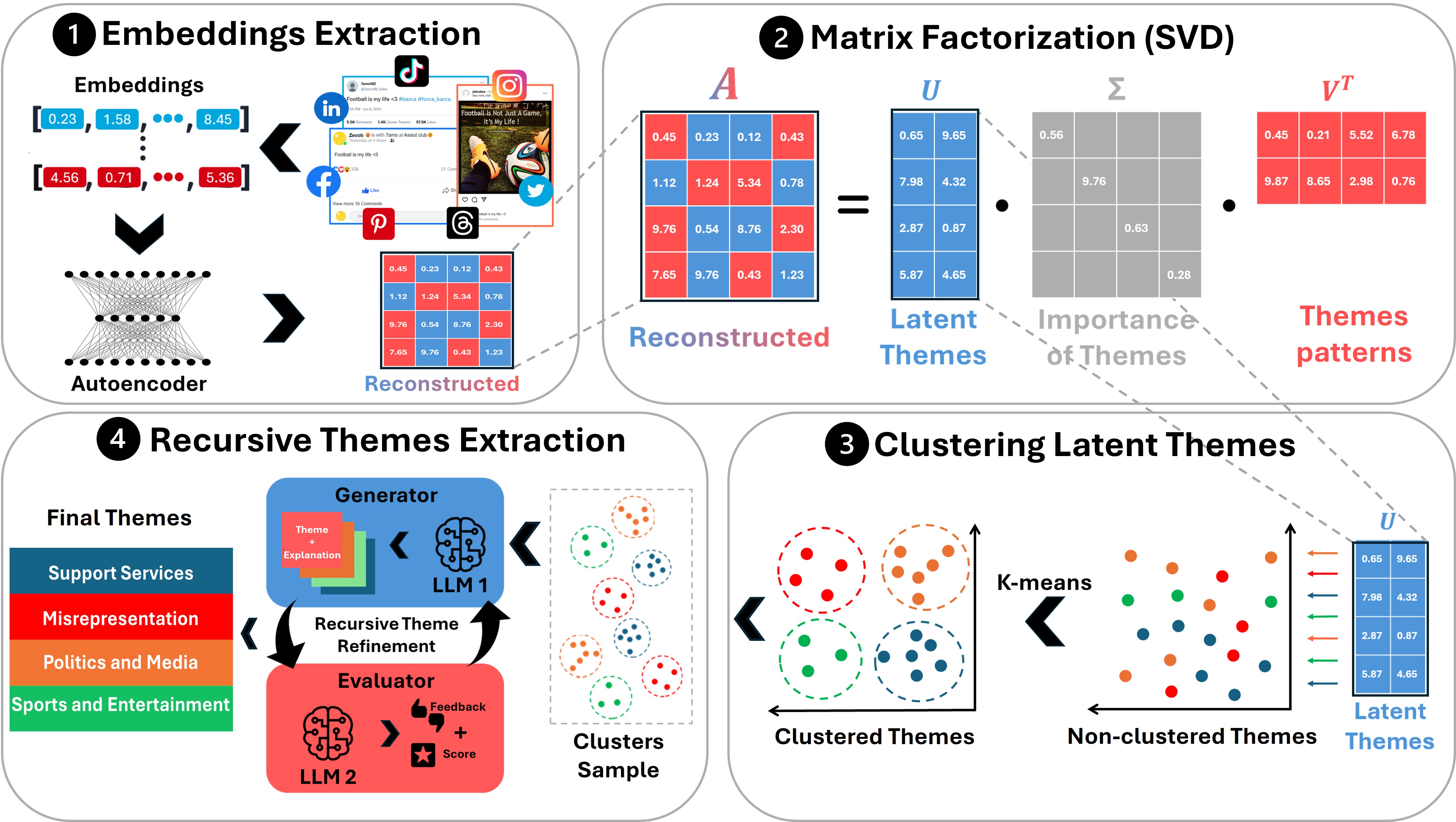}
	\caption{Framework overview of BEYONDWORDS showing posts embedding, compression, matrix factorization, clustering and generative AI based extraction }\label{abstract}
\end{figure}

This research introduces BEYONDWORDS, an agentic generative AI-driven framework for extracting themes from social media. An overview of the methodology is shown in Figure \ref{abstract}

\subsection{Text Tokenization and Embeddings Extraction}

The process of tokenizing and extracting embeddings from text involves converting the raw text into numerical representations that capture both syntactic structure and semantic meaning. Tokenization serves as the initial step, where each tweet is split into a sequence of tokens. Given a tweet \( T \) consisting of \( n \) words \( w_1, w_2, \ldots, w_n \), tokenization can be represented as:

\begin{equation}
    T = \{w_1, w_2, \ldots, w_n\} \rightarrow \{t_1, t_2, \ldots, t_m\}
\end{equation}

where \( t_i \) represents each token after tokenization, and \( m \) is the total number of tokens generated, which depends on the tokenization method and vocabulary size of the embedding model. Most models use a subword-level tokenization, which splits words into smaller units (subwords), capturing even rare or compound words effectively.

After tokenization, each tweet is converted into a fixed-dimensional embedding vector by passing it through a pre-trained language model. These models use large-scale datasets to learn embeddings that encode semantic relationships between words and phrases. For instance, similar words or concepts are represented by vectors that are close to each other in the embedding space, thus capturing semantic similarity. Let \( E \) denote the embedding vector of a tweet, which can be formulated as:

\begin{equation}
    E = Embed\left(t_1, t_2, \ldots, t_m \right)
\end{equation}

where \( Embed \) is the function that processes the token representations and maps them to a high-dimensional space that captures the semantic features of the text.

Three models of different sizes were used to generate embeddings, each capturing semantic features with varying degrees of granularity and complexity. These embeddings have the following characteristics:

\begin{itemize}
    \item \textbf{all-MiniLM-L6-v2} (small) generates embeddings vectors of size 312. Its smaller architecture captures basic semantic relationships, fewer dimensions are less resource-intensive but may miss some subtleties in meaning.
    \item \textbf{bge-base-en-v1.5} (medium) provides embeddings of size 728. This intermediate model balances between efficiency and semantic detail, capturing moderate complexities in sentence structure and meaning.
    \item \textbf{bge-m3} (large) outputs embeddings of size 1024. With a larger number of parameters, it captures nuanced semantic relationships and fine-grained meanings, making it well-suited for tasks requiring high semantic accuracy but comes at a high cost compared to small or medium size representations.
\end{itemize}
Table \ref{tab:embedding_dimensions} summarizes the propoerties of embeddings creation models.

\begin{table}[H]
    \caption{Properties of each model used in text tokenization and embeddings extraction.}
    \centering
    \begin{tabular}{c c c c}
        \hline
        \textbf{Model} & \textbf{Size} & \textbf{Number of Parameters} & \textbf{Dimensions} \\
        \hline
        all-MiniLM-L6-v2 & Small & 23M & 312 \\
        bge-base-en-v1.5 & Medium & 109M & 728 \\
        bge-m3 & Large & 567M & 1024 \\
        \hline
    \end{tabular}
    \label{tab:embedding_dimensions}
\end{table}

The embedding vectors produced by these models capture semantic information by representing words and phrases in a multi-dimensional space, where similar meanings are close together. This first step allows for downstream tasks, such as clustering and classification as explained in subsequent sections.

\subsection{Embeddings Dimensionality Reduction with autoencoder}

To enhance interpretability and reduce computational complexity, embeddings generated from text are further processed through dimensionality reduction using autoencoder. An autoencoder neural network\cite{bank2023autoencoders} consists of two primary components: an encoder \( f_{\theta} \) and a decoder \( g_{\phi} \), parameterized by \( \theta \) and \( \phi \) respectively. Given an input embedding \( E \in \mathbb{R}^d \), the encoder transforms \( E \) through a series of non-linear mappings to a compressed latent representation \( Z \in \mathbb{R}^k \), where \( k < d \). Formally, the encoding process can be described as:

\begin{equation}
Z = f_{\theta}(E) = \sigma \left( W^{(l)} \sigma \left( W^{(l-1)} \cdots \sigma \left( W^{(1)} E + b^{(1)} \right) \cdots + b^{(l-1)} \right) + b^{(l)} \right)
\end{equation}

where \( W^{(i)} \) and \( b^{(i)} \) are the weight matrices and bias vectors for the \( i \)-th layer of the encoder, and \( \sigma \) denotes an activation function (e.g., ReLU or sigmoid). This encoding function learns a transformation that captures the essential structure of \( E \) in the lower-dimensional space \( \mathbb{R}^k \).

The decoder then reconstructs the input embedding \( E \) from the latent representation \( Z \) by applying the inverse transformation, approximating the original embedding through the following mapping:

\begin{equation}
\hat{E} = g_{\phi}(Z) = \sigma \left( W'^{(1)} \sigma \left( W'^{(2)} \cdots \sigma \left( W'^{(m)} Z + b'^{(m)} \right) \cdots + b'^{(2)} \right) + b'^{(1)} \right)
\end{equation}

where \( W'^{(j)} \) and \( b'^{(j)} \) represent the weight matrices and bias vectors of the \( j \)-th layer in the decoder. Here, \( m \) denotes the number of layers in the decoder, and \( \hat{E} \in \mathbb{R}^d \) is the reconstructed embedding that approximates the original \( E \).

The autoencoder neural network is trained by minimizing the reconstruction loss \( L(E, \hat{E}) \), typically formulated as mean squared error (MSE):

\begin{equation}
L(E, \hat{E}) = \frac{1}{d} \sum_{i=1}^{d} \left( E_i - \hat{E}_i \right)^2,
\end{equation}

where \( E_i \) and \( \hat{E}_i \) denote the \( i \)-th components of the original and reconstructed embeddings, respectively. This loss function encourages the model to learn a compact representation that captures the core features of the input embeddings while discarding noise.

Three different compression ratios were explored for the latent dimension \( k \): 1/2, 1/3, and 1/4. Each ratio was assessed to balance the trade-off between dimensionality reduction and reconstruction fidelity, with detailed results discussed in the Results section.

After training, the encoder \( f_{\theta} \) alone is utilized to project high-dimensional embeddings into their corresponding lower-dimensional representations \( Z \), enabling more efficient downstream processing, including Singular Value Decomposition (SVD) for latent theme extraction and clustering with k-means.

\subsection{Matrix Factorization and Clustering}

Matrix factorization using SVD \cite{stewart1993early} was applied to identify underlying themes within the tweet embeddings. Instead of using the original high-dimensional embeddings, the compressed embeddings from the autoencoder neural network were utilized as input. This approach significantly reduces computational cost while preserving essential semantic features, as the dimensionality reduction effectively discards non-informative components.

Given a compressed embedding matrix \( C \in \mathbb{R}^{n \times k} \), where \( n \) is the number of tweet embeddings and \( k \) is the compressed dimension, SVD decomposes \( C \) into three matrices:

\begin{equation}
C = U \Sigma V^T
\end{equation}

where \( U \in \mathbb{R}^{n \times r} \) and \( V \in \mathbb{R}^{k \times r} \) are orthogonal matrices, \( \Sigma \in \mathbb{R}^{r \times r} \) is a diagonal matrix of singular values, and \( r \) represents the rank of \( C \). The columns of \( U \) correspond to the principal components capturing the main themes within the tweet embeddings.

The importance of matrix factorization in this study lies in its ability to distill complex, high-dimensional data into a more manageable form while retaining the most significant features that represent underlying patterns and themes. By decomposing the compressed embeddings, SVD allows for the identification of latent structures that can reveal insights into the thematic content of the social media posts.

To group similar posts based on these themes, \( k \)-means clustering was applied to the reduced representation of the embeddings. This clustering helps in identifying coherent groups of posts, facilitating the extraction of distinct themes. The combination of SVD and \( k \)-means clustering \cite{kodinariya2013review} provides a computationally efficient method to uncover and organize thematic patterns in the data, providing essential component of the proposed approach in this research.
The equation that models this proposed methodology of clustering is expressed as follows:
\begin{equation}
C_{\text{posts}} = \operatorname{k\text{-means}}\left( U, k \right) \quad 
\end{equation}
\begin{equation}
\text{where}  \quad 
U = \operatorname{SVD}_U \left( f_{\theta^*} \left( \operatorname{Embed} \left( \bigcup_{i=1}^N \{t_{i1}, t_{i2}, \dots, t_{im_i}\} \right) \right) \right)
\end{equation}

\begin{equation}
\text{and}  \quad 
\theta^* = \operatorname{argmin}_{\theta} \sum_{i=1}^N L \left( E_i, g_{\phi} \left( f_{\theta}(E_i) \right) \right)
\end{equation}
where the term \(\theta^*\) represents the selection of the optimal compression ratio \(\theta\) that minimize the reconstruction loss \(L(E_i, \hat{E}_i)\) across all embeddings \(E_i\). Here, \(f_{\theta}\) is the encoder function, and \(g_{\phi}\) is the decoder function that reconstructs \(E_i\) from its compressed representation, aiming to minimize the difference between \(E_i\) and its reconstruction \(\hat{E}_i\).

Algorithm \ref{alg:part1} explains the steps performed for clustering the tweets using the proposed approach to prepare them for thematic analysis using LLMs. 

\begin{algorithm}[H]
    \caption{Embedding Extraction and Compression, Matrix Factorization and Clustering}
    \label{alg:part1}
    \nolinenumbers
    \begin{algorithmic}
        \State \textbf{Input:} Tweets dataset $T$, Compression ratios $CR$, Number of clusters $k$
        \State \textbf{Output:} Clustered posts $C_{posts}$

        \State \text{1. Text Tokenization:}
        \State \quad For each tweet \( T_i \in T \):
        \State \quad \quad $T_i = \{w_1, w_2, \ldots, w_n\} \rightarrow \{t_1, t_2, \ldots, t_m\}$

        \State \text{2. Embedding Extraction:}
        \For{each model in \{X, Y, Z\}}
            \State \quad $E_i \gets \operatorname{Embed}(t_1, t_2, \ldots, t_m)$
        \EndFor

        \State \text{3. Dimensionality Reduction with Autoencoders:}
        \For{each compression ratio $\theta \in CR$}
            \State \quad $Z_i^{(\theta)} = f_{\theta}(E_i)$ \quad \text{(Encode)}
            \State \quad $\hat{E}_i^{(\theta)} = g_{\phi_\theta}(Z_i^{(r)})$ \quad \text{(Decode)}
            \State \quad Calculate $L(E_i, \hat{E}_i^{(r)})$
        \EndFor
        \State \quad Select $\theta^* = \operatorname{argmin}_{\theta \in CR} \sum_{i} L(E_i, \hat{E}_i^{(r)})$
        \State \quad Set $Z_i = Z_i^{(\theta^*)}$ for each $i$

        \State \text{4. Matrix Factorization (SVD):}
        \State \quad $C \gets \text{Compressed Embedding Matrix for selected } \theta^*$
        \State \quad Decompose $C$ using SVD: $C = U \Sigma V^T$

        \State \text{5. Clustering:}
        \State \quad $C_{posts} \gets k\text{-means}(U, k)$ \quad \text{(Cluster posts)}
        
    \end{algorithmic}
\end{algorithm}

\subsection{Generative AI for Themes Extraction}
To effectively manage the analysis of social media posts, a two-step sampling strategy was employed. It is based on the Cochran formula to determine an appropriate sample size from each cluster. This approach allowed conducting a focused thematic analysis while adhering to the limitations of generative AI regarding context windows and computational costs.
The Cochran formula for sample size determination is given by:

\begin{equation}
n = \frac{Z^2 \cdot p \cdot (1 - p)}{e^2}
\end{equation}
Where: \(n\) is the sample size, \(Z\) is the Z-score corresponding to the desired confidence level (e.g., 1.64 for a 90\% confidence level), \(p\) is the estimated proportion of the population (we use \(0.5\) for maximum sample size), \(e\) is the margin of error (the desired level of precision).

Using this formula, the sample size \(n\) for each cluster was calculated. For a 90\% confidence the final sample size is determined to be 68. This statistically representative sample enables the model to apply the thematic extraction methodology efficiently without processing the entire batch of posts.
Following the determination of the sample size, the selection was further improved by employing the silhouette score to identify the top \(n\) texts with the highest cohesion and separation from other clusters. The silhouette score is defined as:

\begin{equation}
s(i) = \frac{b(i) - a(i)}{\max\{a(i), b(i)\}}
\label{silhouette}
\end{equation}
Where: \(s(i)\) is the silhouette score for the \(i\)-th data point, \(a(i)\) is the average distance between the \(i\)-th data point and all other points in the same cluster, \(b(i)\) is the average distance between the \(i\)-th data point and all points in the nearest cluster.

By calculating the silhouette score for all posts within each cluster, the texts selected are the highest scorers, indicating that they are well-clustered and representative of their themes. 

\subsubsection{CoT Strategy}

The CoT strategy serves as a pivotal component of the proposed methodology, guiding the generative process through structured and sequential tasks. The process can be detailed as follows:

\begin{enumerate}
    \item The LLM is prompted to identify significant keywords and phrases within the tweets, focusing on content that reflects important topics and sentiments.
    \item The extracted keywords are organized into coherent groups based on common themes, topics, or ideas, enhancing clarity and coherence.
    \item For each category, the LLM work as a specialized agent in thematic extraction, synthesizes high-level themes, providing concise descriptions that encapsulate the essence of the discussions in the tweets.
\end{enumerate}

\subsubsection{Recursive Theme Refinement via LLM Feedback Mechanism}

To further refine the theme extraction process, a recursive feedback mechanism involving a second agent LLM is implemented. This agent acts as a grading system, following these steps:

\begin{enumerate}
    \item The secondary LLM evaluates the themes generated by the primary LLM against predefined quality thresholds.
    \item If the themes do not meet the established criteria, feedback is relayed back to the primary LLM.
    \item The primary LLM utilizes the feedback (score + comment) to reevaluate the extraction process, revisiting the previous steps as necessary.
    \item This cycle of evaluation and adjustment continues, improving the quality of theme extraction with each iteration until either the threshold score or the maximum number of iterations are reached.
\end{enumerate}
    
This iterative refinement mimics the Generative Adversarial Networks (GANs) paradigm \cite{goodfellow2020generative}, where a generator produces outputs while a discriminator assesses their quality. In this context, the generator is the primary LLM that generates thematic outputs, while the discriminator is the secondary LLM that critiques these outputs. Since the discriminator LLM provides the score and feedback as part of a sentence, a mechanism is needed to extract only the score and accurate feedback. This is essential to isolate the components required to decide whether the extracted theme is good enough by comparing the score with a threshold, or to pass only these two elements to the first step to redo the extraction process, taking into consideration the new score and feedback from the discriminator. To serve this purpose, we use an LLM-based entity extractor model \cite{ghali2024gamedx}. This discourse between LLMs ensures continuous improvement in theme extraction, allowing the methodology to adapt and enhance its performance with each iteration. By combining structured prompting with feedback-driven refinement. 

The following equation describes the recursive refinement strategy proposed by modeling the theme extraction process for cluster \( k \), where \( T_{\text{themes}, k} \) is the final set of themes. It uses the indicator function \( \mathbb{I} \) to choose between refining the initial themes, \( T_{\text{initial}} \), or using them directly, based on the evaluation score. If the score is below the threshold \( Q \), the themes are refined using feedback through \( \mathcal{R}(T, \text{feedback}) \), otherwise, the initial themes are selected. The scoring function \( \mathcal{S}(\text{score}(T)) \) evaluates the quality of the themes, while \( \mathcal{M}_1 \) and \( \mathcal{M}_2 \) represent the language models used for theme extraction and evaluation, respectively.
\begin{equation}
T_{\text{themes}, k} = 
\arg \max_{T} \left[ \mathbb{I}_{\left( \mathcal{S}(\text{score}(T)) < Q \right)} \cdot \mathcal{R}(T, \text{feedback}) + \mathbb{I}_{\left( \mathcal{S}(\text{score}(T)) \geq Q \right)} \cdot T_{\text{initial}} \right]
\end{equation}

\(
\text{where} \quad 
T_{\text{initial}} = \mathcal{M}_1 \left( C_{\text{posts}, k} \right)
\quad \text{and} \quad
\left( \text{score}, \text{feedback} \right) = \mathcal{M}_2 \left( T_{\text{initial}} \right)
\)

Algorithm \ref{alg:genai} summarizes in details the steps to extract the themes:
\begin{algorithm}[H]
    \caption{Agentic CoT Model for Theme Extraction}
    \label{alg:genai}
    \nolinenumbers
    \begin{algorithmic}
        \State \textbf{Input:} Clustered posts $C_{posts,k}$
        \State \textbf{Output:} Extracted themes for each cluster $k$: $T_{themes,k}$
        
        \For{each cluster $k$ in $C_{posts,k}$}
            \State \text{1. Sample Selection:}
            \State \quad Calculate sample size $n$ using Cochran's formula
            \State \quad Sample $n$ posts from cluster $k$

            \State \text{2. Initial Theme Extraction:}
            \State \quad $T_{initial} \gets \text{LLM}_1.extract\_themes(C_{posts,k})$

            \State \text{3. Quality Evaluation:}
            \State \quad $score, feedback \gets \text{LLM}_2.evaluate\_and\_extract(T_{initial})$
            
            \If{$score < Q$}
                \State \text{4. Feedback and Refinement:}
                \State \quad $T_{refined} \gets \text{LLM}_1.refine\_themes(T_{initial}, feedback)$
                \State \quad $T_{themes,k} \gets T_{refined}$
            \Else
                \State \quad $T_{themes,k} \gets T_{initial}$
            \EndIf
        \EndFor
        
        \State \text{5. Output Extracted Themes:} $T_{themes,k}$ for each cluster $k$
    \end{algorithmic}
\end{algorithm}

\section{Autism Case Study}

\subsection{Dataset Overview}

\begin{figure}[H]
	\centering
	\includegraphics[width=\textwidth]{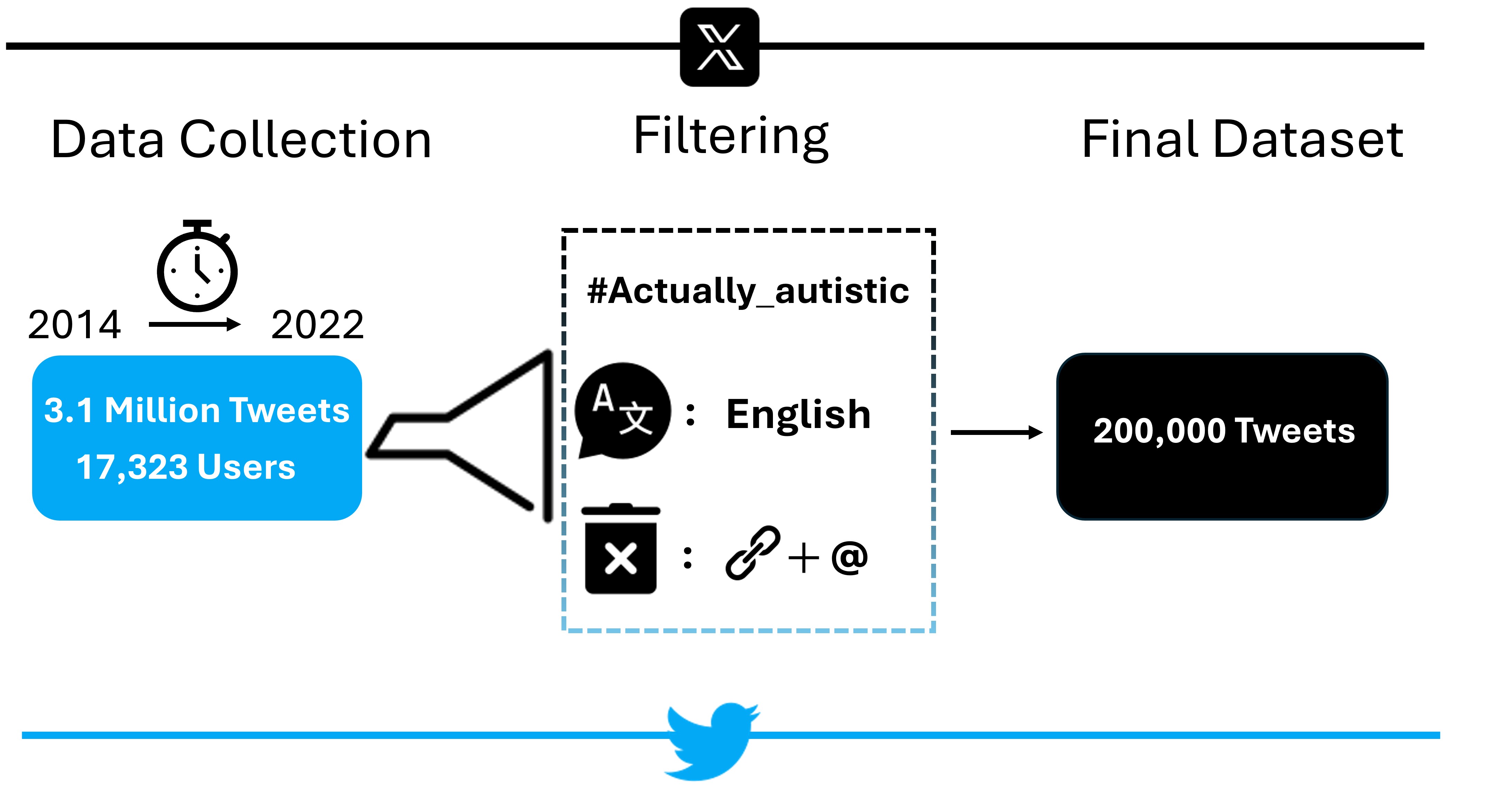}
	\caption{Overview of the dataset preprocessing used in this research }\label{datapreprocessing}
\end{figure}

The dataset used for this study comprises a selection of tweets from individuals who identify as autistic, using the hashtag \#actuallyautistic to share their experiences, insights, and perspectives. This hashtag represents a movement within the autism community that seeks to amplify the voices of autistic individuals offering a more personal and direct narrative on living with autism. 
The original dataset was gathered using \texttt{snscrape}, a Python-based scraping tool, to extract tweets from X(Twitter previously) between January 2014 and December 2022. The dataset includes tweets from individuals who self-identified as autistic through keywords such as “autism,” “autistic,” or “neurodiverse” in their profiles, usernames or tweets. This initial dataset consists of approximately 3.1 million tweets from 17,323 unique users, with additional metadata on usernames, account creation dates, and other relevant features.
Figure \ref{datapreprocessing} presents an overview of the preprocessing process.

\subsection{Dataset Preprocessing}
From the original dataset \cite{jaiswal2023actuallyautistic}, only tweets that contained \#actuallyautistic hashtag with its different textual variations were retained to ensure relevance to autistic individuals. Additionally, tweets were filtered to include only those written in English, resulting in a subset of approximately 200,000 tweets for further analysis. To prepare the data for processing, a text-cleaning script was applied to each tweet, removing URLs, mentions, hashtags, and special characters. This preprocessing step created a refined dataset that maintains linguistic consistency and readability for subsequent analysis.

\section{Experiments Results and Discussion}
\subsection{Dimensionality Reduction Analysis}
This section presents the findings from experiments conducted on autoencoder neural network models designed for dimensionality reduction of text embeddings. The primary goal is to evaluate the performance of three distinct model architectures—small, medium, and large—across varying levels of compression. Specifically, the analysis focuses on how each model performs with three different compression ratios: 1/2, 1/3, and 1/4 of the original embedding dimensionality. The ultimate goal is to identify the model that minimizes information loss, ensuring the highest possible accuracy for subsequent SVD and clustering using the minimum computations possible.

\begin{figure}[H]
	\centering
	\includegraphics[width=\textwidth]{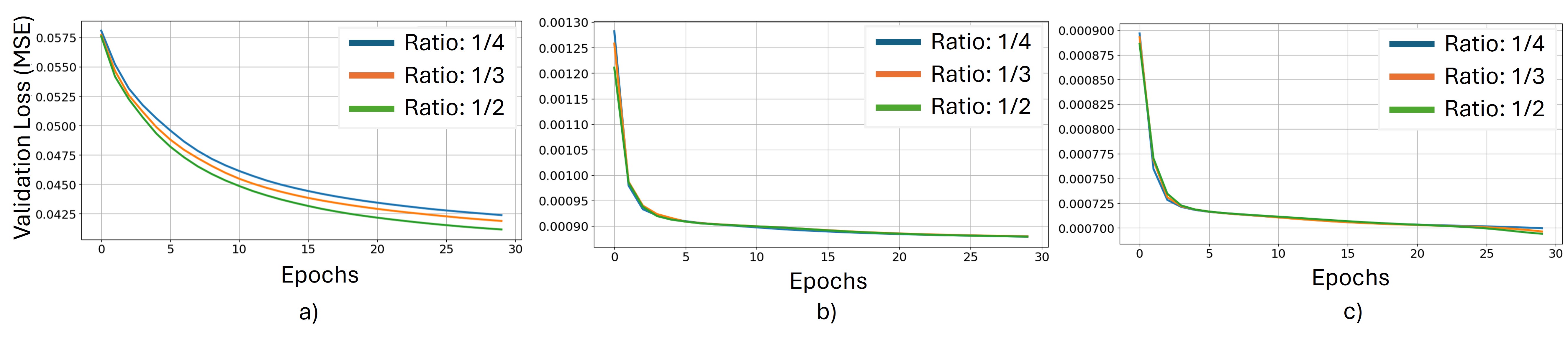}
	\caption{Autoencoder loss plot for a) small, b) medium, and c) large embedding models}\label{loss}
\end{figure}
Figure \ref{loss} presents the validation loss across training epochs for the autoencoder model. For the small embedding size (384 dimensions), as shown in Figure \ref{loss}a, the 1/2 compression ratio achieves the lowest validation loss, indicating that halving the dimensionality retains the essential features for accurate reconstruction, balancing information preservation and reduced computational complexity. For the medium embedding size (768 dimensions), highlighted in Figure \ref{loss}b, the 1/4 compression ratio is optimal, suggesting that medium-sized embeddings tolerate higher compression while still enabling precise reconstruction. This indicates that the medium model is more resilient to dimensionality reduction, allowing for more efficient computation in subsequent SVD and clustering steps. For the large embedding size (1024 dimensions), as shown in Figure \ref{loss}c, the best performance is again at the 1/2 compression ratio, reflecting that halving the dimensionality remains optimal for capturing necessary data patterns despite the larger embedding size. Overall, the analysis highlights the autoencoder’s ability to retain critical information across all embedding sizes and compression ratios, as evidenced by the convergence of validation loss to low values. The preservation of embedding structure post-compression is crucial for accurate clustering with k-means, enabling efficient theme extraction with minimal information loss.

\subsection{Matrix Factorization and Clustering}
\begin{figure}[H]
	\centering
	\includegraphics[width=\textwidth]{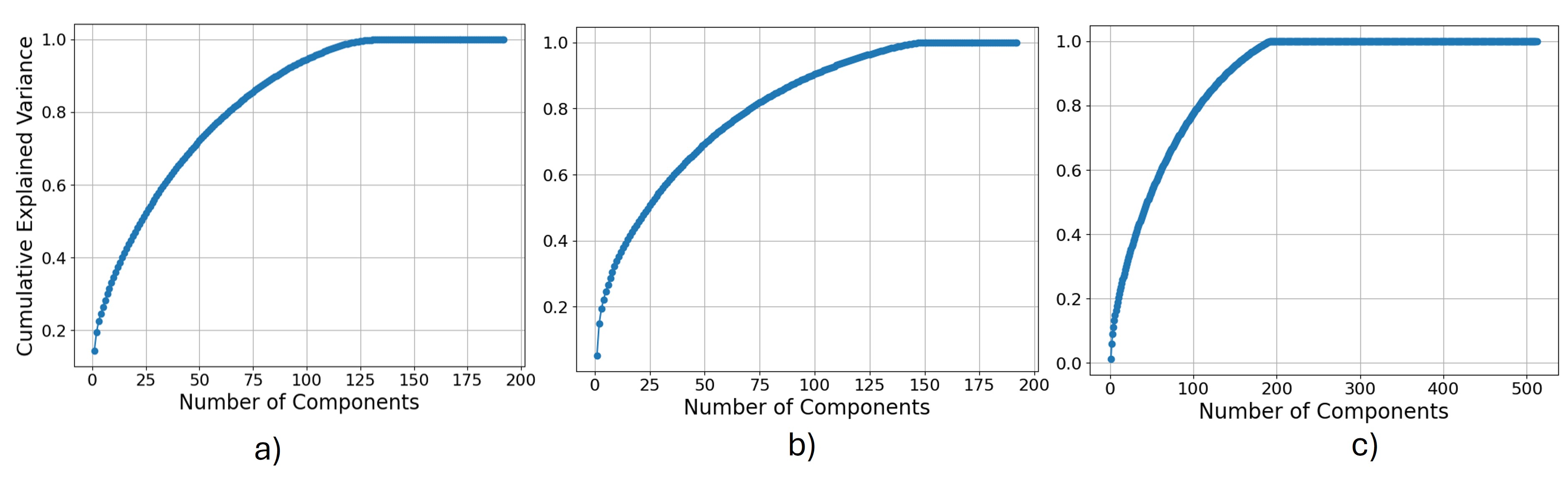}
	\caption{SVD plot for a) small, b) medium, and c) large embedding models}\label{svdcombined}
 
\end{figure}

The results of the SVD analysis, illustrated in Figure \ref{svdcombined}, demonstrate the effectiveness of the proposed dimensionality reduction methodology for embeddings of varying sizes (small, medium, and large). Following the initial compression achieved by autoencoders, SVD was applied to further reduce the embeddings' dimensionality, facilitating efficient clustering and theme extraction. This step is particularly crucial in the methodology, as matrix factorization not only enhances interpretability but also optimizes computational resources, making it well-suited for large-scale social media posts analysis.
In Figure \ref{svdcombined}a, which corresponds to the small-size embeddings, the cumulative explained variance approaches 90\% with approximately 100 components. This finding suggests that SVD effectively captures the majority of the data’s variance, while significantly reducing the dimensional space. 
For medium-size embeddings, shown in Figure \ref{svdcombined}b, a similar pattern emerges. The cumulative explained variance also reaches close to 90\% with fewer than 100 components. This consistency across both small and medium embeddings indicates that SVD is highly effective in distilling essential features, even as the embedding size increases.
Figure \ref{svdcombined}c presents the results for large-size embeddings, where approximately 150 components are required to capture a comparable level of variance. This increase in required components is expected, as larger embeddings inherently contain more complex information. Despite this, the curve stabilizes, confirming that SVD provides a robust means of reducing even high-dimensional embeddings to a manageable form, preserving critical information for clustering.


\subsection{Clustering (combine all 6 pictures in one big picture)}

\begin{figure}[H]
	\centering
	\includegraphics[width=\textwidth]{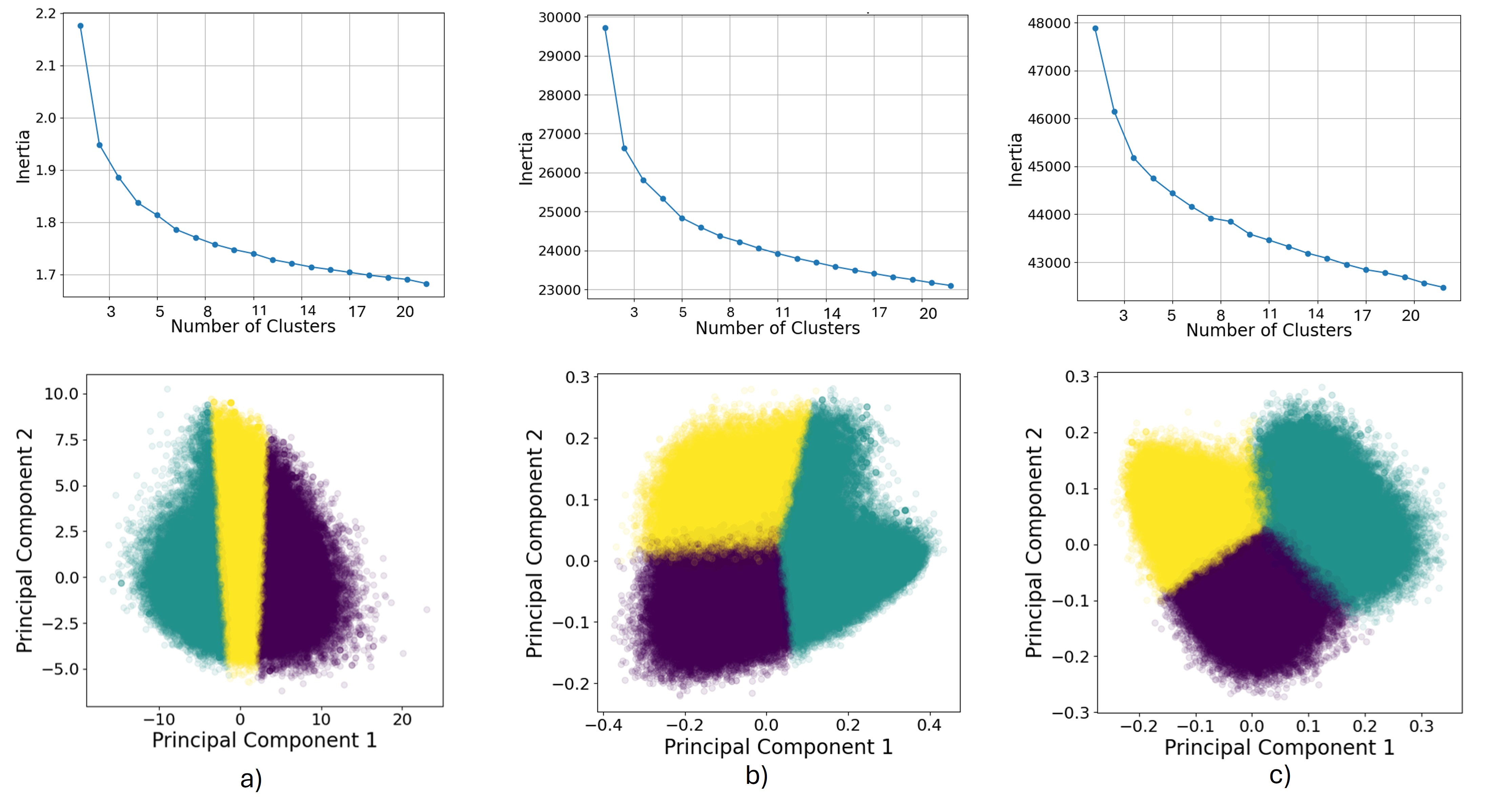}
	\caption{Elbow and k-means themes clusters plot for a) small, b) medium, and c) large embedding models}\label{combinedclusters}
\end{figure}

The elbow method (Figure \ref{combinedclusters}) was used to determine the optimal number of clusters for latent themes derived via SVD from compressed embeddings. For all embedding sizes (Figures \ref{combinedclusters}a, \ref{combinedclusters}b, \ref{combinedclusters}c), the elbow point consistently appears at 3 clusters suggesting that additional clusters would contribute minimal new information, confirming that 3 clusters capture the primary themes in the data.
The clustering results demonstrate the robustness and scalability of the proposed methodology. Across all embedding sizes, three distinct clusters are consistently formed, each reflecting well-separated themes within the social media data. 
In Figure \ref{combinedclusters}a, the small-sized embeddings display clear separation among the clusters, with each color representing a unique latent theme. The compact arrangement of points within each cluster suggests that the themes identified are coherent and well-differentiated even with a limited embedding dimensionality. Figure \ref{combinedclusters}b, representing medium-sized embeddings, shows a similar pattern, with clusters that retain distinct boundaries and minimal overlap. This demonstrates that the clustering method is adaptable, producing stable clusters as the embedding dimensionality increases.
Figure \ref{combinedclusters}c, showing results for the large-sized embeddings, presents a particularly compelling case for the methodology's effectiveness. The clusters remain well-defined, with minimal inter-cluster overlap, despite the increase in data complexity and embedding dimensionality. The large embedding size allows for capturing more nuanced details in the data, yet the clusters continue to exhibit clear separation, indicating that the approach is capable of handling high-dimensional representations while preserving theme clarity.
The consistency of cluster shapes and separation across embedding sizes reinforces the suitability of this methodology for identifying latent themes in social media data. The SVD step plays a pivotal role in this outcome by reducing dimensionality while retaining essential information, enabling k-means clustering to form coherent clusters that are robust to changes in embedding size.



To accurately measure the quality of the clustering results after applying the proposed methodology, and compare the effectiveness of using autoencoder neural network for dimensionality reduction in text embedding compression, it was essential to conduct an analysis with and without autoencoder-based compression across different embedding sizes. This analysis allows for a deeper understanding of how compression impacts clustering quality, which is crucial when applying subsequent dimensionality reduction techniques like SVD for latent theme extraction and clustering. To evaluate the clustering performance, three metrics were considered:
\textbf{Calinski-Harabasz Index (CH Index):} Measures the ratio of the sum of between-cluster dispersion to within-cluster dispersion. Higher values indicate better-defined clusters.
   \begin{equation}
   \text{CH Index} = \frac{\text{Tr}(B_k)}{\text{Tr}(W_k)} \cdot \frac{N - k}{k - 1}
   \end{equation}
   where \( \text{Tr}(B_k) \) is the trace of the between-cluster dispersion matrix, \( \text{Tr}(W_k) \) is the trace of the within-cluster dispersion matrix, \( N \) is the total number of points, and \( k \) is the number of clusters.
\newline
\newline
\textbf{Davies-Bouldin Index (DB Index):} Measures the average "similarity" ratio of within-cluster distances to between-cluster distances. Lower values indicate better clustering quality.
   \begin{equation}
   \text{DB Index} = \frac{1}{k} \sum_{i=1}^{k} \max_{j \neq i} \left( \frac{\sigma_i + \sigma_j}{d(c_i, c_j)} \right)
   \end{equation}
   where \( \sigma_i \) and \( \sigma_j \) are the within-cluster distances for clusters \( i \) and \( j \), respectively, and \( d(c_i, c_j) \) is the distance between cluster centers \( c_i \) and \( c_j \).
   \newline
   \newline
\textbf{Silhouette Score:} Reflects the compactness and separation of clusters. It ranges from -1 to 1, where higher values indicate well-separated clusters as explained in details in Equation \ref{silhouette}.
\newline
The tables below summarizes the clustering metrics results for different embedding sizes with and without autoencoder-based compression.

\begin{table}[H]
\centering
\small
\caption{Clustering metrics for small size embeddings}
\begin{tabular}{lcc}
\hline
\textbf{Metrics} & \textbf{With Autoencoder} & \textbf{Without Autoencoder} \\
                 & \textbf{Compression}      & \textbf{Compression} \\
\hline
CH Index & 24991 & 5161 \\
DB Index & 2.93 & 8.22 \\
Silhouette Score & 0.04 & 0.04 \\
\hline
\end{tabular}
\label{table:small}
\end{table}

\begin{table}[H]
\centering
\small
\caption{Clustering metrics for medium size embeddings}
\begin{tabular}{lcc}
\hline
\textbf{Metrics} & \textbf{With Autoencoder} & \textbf{Without Autoencoder} \\
                 & \textbf{Compression}      & \textbf{Compression} \\
\hline
CH Index & 11806 & 11879 \\
DB Index & 4.01 & 3.80 \\
Silhouette Score & 0.06 & 0.06 \\
\hline
\end{tabular}
\label{table:medium}
\end{table}

\begin{table}[H]
\centering
\small
\caption{Clustering metrics for large size embeddings}
\begin{tabular}{lcc}
\hline
\textbf{Metrics} & \textbf{With Autoencoder} & \textbf{Without Autoencoder} \\
                 & \textbf{Compression}      & \textbf{Compression} \\
\hline
CH Index & 366243 & 6235 \\
DB Index & 0.62 & 5.22 \\
Silhouette Score & 0.48 & 0.04 \\
\hline
\end{tabular}
\label{table:large}
\end{table}

The clustering metrics presented in Tables \ref{table:small}, \ref{table:medium}, and \ref{table:large} serve as the final quality assessment of the methodology for extracting themes from text embeddings. This analysis follows the dimensionality reduction via autoencoder, SVD for latent theme extraction, and clustering with k-means. The results validate the effectiveness of this multi-step approach in capturing meaningful structure within the data.
The high CH Index values and low DB Index values achieved, particularly in the configurations with autoencoder neural network compression, demonstrate the ability of this methodology to create well-separated and compact clusters, indicative of high clustering quality. For instance, in the large embedding size, the CH Index reaches 366243 with a DB Index of 0.62 when autoencoder neural network are used, signaling exceptionally distinct clusters. This trend is consistent across embedding sizes, with the small embedding size achieving a CH Index of 24991 and a DB Index of 2.93, and the medium size showing similarly favorable metrics. 
Moreover, the silhouette scores across the tests further confirm the consistency and stability of the clustering structure. These metrics collectively indicate that the clusters generated are not only compact and well-separated but also robust in representing latent themes within the data as evidenced by this final quality assurance test. 

\subsection{Generative AI Theme Extraction Analysis}
In the thematic extraction phase, large embedding model's themes clusters were used as it achieved the lowest loss value and highest clustering quality across two out of three evaluation metrics. GPT -4o mini model was used for both the primary extractor (LLM1) and evaluator (LLM2) roles as it is known for its lightweight design, per token cost effectiveness and strong performance across major evaluation benchmarks \cite{openai2024gpt4ocard}. This analysis section presents both word cloud visualizations of keywords and Sankey diagrams illustrating the connections between keywords and their coherent groups, providing a comprehensive view of the relationships identified to further examine and validate the final themes extracted.

\begin{figure}[H]
	\centering
	\includegraphics[width=\textwidth]{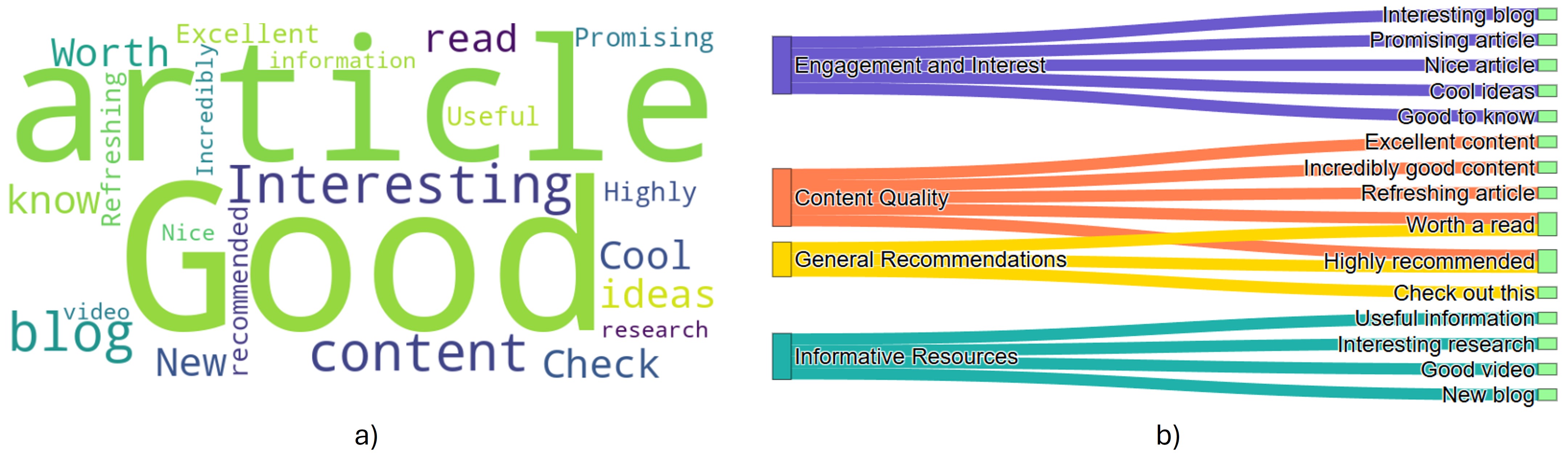}
	\caption{LLMs based thematic analysis for theme 1: a) word cloud of keywords extracted by LLM1, b) Sankey diagram of keywords extracted and their coherent groups clustering}\label{theme1}
\end{figure}

\begin{figure}[H]
	\centering
	\includegraphics[width=\textwidth]{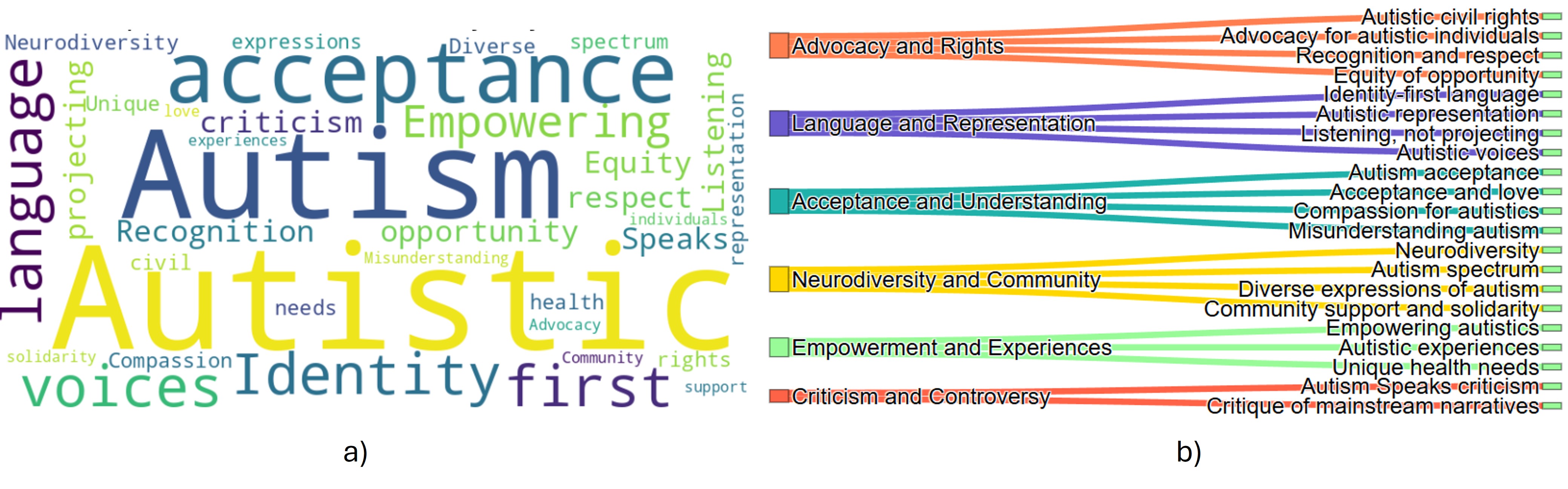}
	\caption{LLMs based thematic analysis for theme 2: a) word cloud of keywords extracted by LLM1, b) Sankey diagram of keywords extracted and their coherent groups clustering}\label{theme2}
\end{figure}

\begin{figure}[H]
	\centering
	\includegraphics[width=\textwidth]{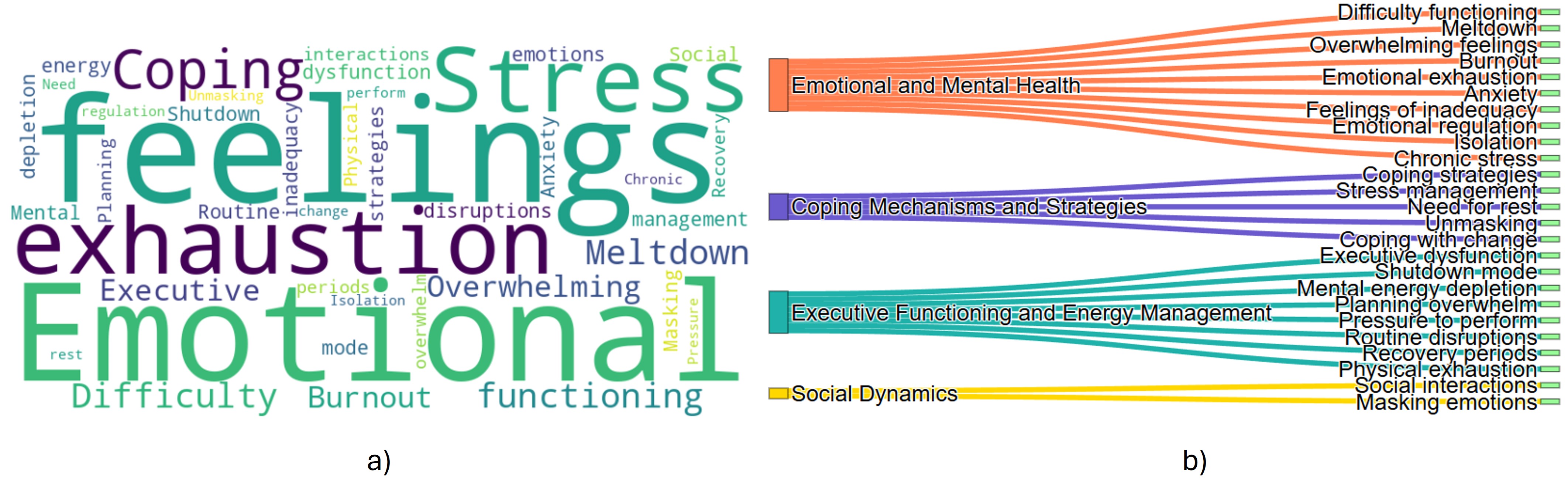}
	\caption{LLMs based thematic analysis for theme 3: a) word cloud of keywords extracted by LLM1, b) Sankey diagram of keywords extracted and their coherent groups clustering}\label{theme3}
\end{figure}

Through the structured CoT and recursive theme refinement strategy explained in details in methodology section, the LLM1 first identified significant keywords and phrases in the tweets text, focusing on elements that reflect key topics and sentiments. These keywords were then organized into multiple coherent groups for each cluster yielding at the end the final high-level themes:

\begin{itemize}
    \item \textbf{Social media content quality and engagement:} This theme contains keywords related to the quality and engagement potential of social media content, such as "Excellent content," "Highly recommended," and "Promising article." The associated Sankey diagram (Figure \ref{theme1}b) illustrates the connections between these engagement-oriented terms, while the word cloud (Figure \ref{theme1}a) highlights the prominence of keywords within this theme.
    
    \item \textbf{Advocacy for autistic rights and acceptance:} This theme emphasizes the importance of representation, empowerment, and critiques of dominant autism narratives. Keywords like "Autistic civil rights," "Neurodiversity," and "Acceptance and love" underscore the advocacy focus within this group. Figure~\ref{theme2}b provides the Sankey diagram showing how terms in this theme are interrelated, and Figure~\ref{theme2}a displays the word cloud, showcasing the relevance of advocacy-related terms.

    \item \textbf{Mental health and well-being:} This theme captures mental health challenges and coping strategies, with keywords like "Burnout," "Executive dysfunction," "Coping strategies," and "Isolation." The Sankey diagram (Figure~\ref{theme3}b) shows connections between terms related to mental health, while Figure~\ref{theme3}a displays a word cloud of keywords within this theme, underscoring the emphasis on managing mental and emotional challenges.
\end{itemize}

The themes revealed by LLM1 were iteratively evaluated by the LLM2 until they reached the predefined quality acceptance criteria. If the themes did not meet these criteria, feedback was provided to LLM1 for adjustments. This recursive feedback loop significantly enhanced the clarity and coherence of the identified themes, allowing for an even better and accurate thematic extraction as evidenced by the final themes above.

\subsection{Discussion}

\subsubsection{Thematic Study Insights}
The thematic analysis provided valuable insights across the three primary areas, highlighting interesting discussions and sentiments within the social media content of autistic people.

The first extracted theme \textbf{social media content quality and engagement} reveals high engagement indicators such as "Excellent content," "Promising article," and "Highly recommended," indicating a strong focus on content quality. This underscores the value placed on credible, high-quality resources in social media conversations and what it means to autistic people.
    
The second theme \textbf{advocacy for autistic rights and acceptance} with keywords such as "Autistic civil rights," "Neurodiversity," and "Acceptance and love" reflect the community's emphasis on representation and self-determined identity. This theme highlights the role of social media in autistic advocacy, providing a platform to advocate for acceptance, critique organizations (e.g., “Autism Speaks criticism”), and promote neurodiversity.
    
The last extracted theme \textbf{mental health and well-being} groups keywords around mental health challenges and coping strategies, such as "Burnout," "Executive dysfunction," and "Coping strategies." The keywords highlight the community's struggles with mental and emotional well-being, as well as the importance of resilience and coping mechanisms, particularly in the context of neurodiverse experiences.

\subsubsection{Implications for Autism Advocacy and Informed Decision-Making}
The identified themes have implications for both advocacy and informed decision-making. For example, the second theme reveals community-driven discussions about identity-first language and critiques of existing narratives, guiding policymakers and advocacy groups to better align their communication and policies with community values.

In addition, the third theme underscores the need for targeted mental health resources for autistic individuals, pointing to the importance of tailored interventions for this population. These insights support the goal of using data-driven themes to foster inclusivity and improve access to relevant resources without neglecting the fact that they value quality content on social media as depicted by the first theme.

\section{Conclusion}
This paper presents BEYONDWORDS, a novel methodology for extracting latent themes from social media posts, with a specific focus on content related to autistic individuals. The approach integrates embeddings, dimensionality reduction, SVD for theme extraction, K-means clustering, and an agentic generative AI model with iterative feedback. This pipeline successfully identified three primary themes: \textbf{social media content quality and engagement}, \textbf{advocacy for autistic rights and acceptance}, and \textbf{mental health and well-being}. The thematic analysis provides valuable insights, revealing high engagement with quality content, a strong emphasis on autistic rights and acceptance, and the importance of mental health and coping strategies within the autistic community. While the methodology offers a comprehensive and scalable approach to theme extraction, it is not without limitations. The complexity of the multistep process and the dependency on high-quality embedding models are notable challenges. Additionally, potential biases in the AI models and the sensitivity of SVD and K-means to parameter tuning require careful consideration.
Future work will focus on addressing these limitations by developing techniques to mitigate biases, exploring real-time processing capabilities, and incorporating user feedback mechanisms to further refine the accuracy and applicability of the approach. These enhancements will make the methodology a robust tool for social media analysis, particularly in understanding and supporting the autistic community.




\end{document}